\documentclass[sigconf, screen, table]{acmart}

\AtBeginDocument{%
  \providecommand\BibTeX{{%
    \normalfont B\kern-0.5em{\scshape i\kern-0.25em b}\kern-0.8em\TeX}}}

\usepackage{bm}
\usepackage{multirow}
\usepackage{balance}

\def\eg{\emph{e.g}.} 
\def\ie{\emph{i.e}.}
\def\wrt{\emph{w.r.t}.}
\def\etc{\emph{etc}}

\definecolor{orange}{rgb}{1, 0.85, 0.7}
\definecolor{tablered}{rgb}{1, 0.7, 0.7}

\copyrightyear{2023}
\acmYear{2023}
\setcopyright{acmlicensed}
\acmConference[MM '23]{Proceedings of the 31st ACM International Conference on Multimedia}{October 29-November 3, 2023}{Ottawa, ON, Canada}
\acmBooktitle{Proceedings of the 31st ACM International Conference on Multimedia (MM '23), October 29-November 3, 2023, Ottawa, ON, Canada}
\acmPrice{15.00}
\acmDOI{10.1145/3581783.3611857}
\acmISBN{979-8-4007-0108-5/23/10}

\begin{document}

\title{Mirror-NeRF: Learning Neural Radiance Fields for Mirrors with Whitted-Style Ray Tracing}

\author{Junyi Zeng}
\authornote{Junyi Zeng and Chong Bao contributed equally to this research. The authors from Zhejiang University are also affiliated with the State Key Lab of CAD\&CG.}
\affiliation{
  \institution{Zhejiang University}
  \city{Hangzhou}
  \country{China}
}
\email{zengjunyi@zju.edu.cn}

\author{Chong Bao}
\authornotemark[1]
\affiliation{
  \institution{Zhejiang University}
  \city{Hangzhou}
  \country{China}
}
\email{chongbao@zju.edu.cn}

\author{Rui Chen}
\affiliation{
  \institution{Zhejiang University}
  \city{Hangzhou}
  \country{China}
}
\email{22221111@zju.edu.cn}

\author{Zilong Dong}
\affiliation{
  \institution{Alibaba Group}
  \city{Hangzhou}
  \country{China}
}
\email{list.dzl@alibaba-inc.com}

\author{Guofeng Zhang}
\affiliation{
  \institution{Zhejiang University}
  \city{Hangzhou}
  \country{China}
}
\email{zhangguofeng@zju.edu.cn}

\author{Hujun Bao}
\affiliation{
  \institution{Zhejiang University}
  \city{Hangzhou}
  \country{China}
}
\email{bao@cad.zju.edu.cn}

\author{Zhaopeng Cui}
\affiliation{
  \institution{Zhejiang University}
  \city{Hangzhou}
  \country{China}
}
\email{zhpcui@zju.edu.cn}
\authornote{Corresponding author: Zhaopeng Cui.}

\renewcommand{\shortauthors}{Junyi Zeng and Chong Bao, et al.}

%%%%%%%%% ABSTRACT
\begin{abstract}
   Recently, Neural Radiance Fields (NeRF) has exhibited significant success in novel view synthesis, surface reconstruction, \etc.
   However, since no physical reflection is considered in its rendering pipeline, NeRF mistakes the reflection in the mirror as a separate virtual scene, leading to the inaccurate reconstruction of the mirror and multi-view inconsistent reflections in the mirror.
   In this paper, we present a novel neural rendering framework, named Mirror-NeRF, which is able to learn accurate geometry and reflection of the mirror and support various scene manipulation applications with mirrors, such as adding new objects or mirrors into the scene and synthesizing the reflections of these new objects in mirrors, controlling mirror roughness, \etc.
   To achieve this goal,  we propose a unified radiance field by introducing the reflection probability and tracing rays following the light transport model of Whitted Ray Tracing, and also develop several techniques to facilitate the learning process.
   Experiments and comparisons on both synthetic and real datasets demonstrate the superiority of our method.
   The code and supplementary material are available on the project webpage: 
   \urlstyle{tt}{\url{https://zju3dv.github.io/Mirror-NeRF/}.}
\end{abstract}




\begin{teaserfigure}
  \vspace{-1em}
  \includegraphics[width=\textwidth]{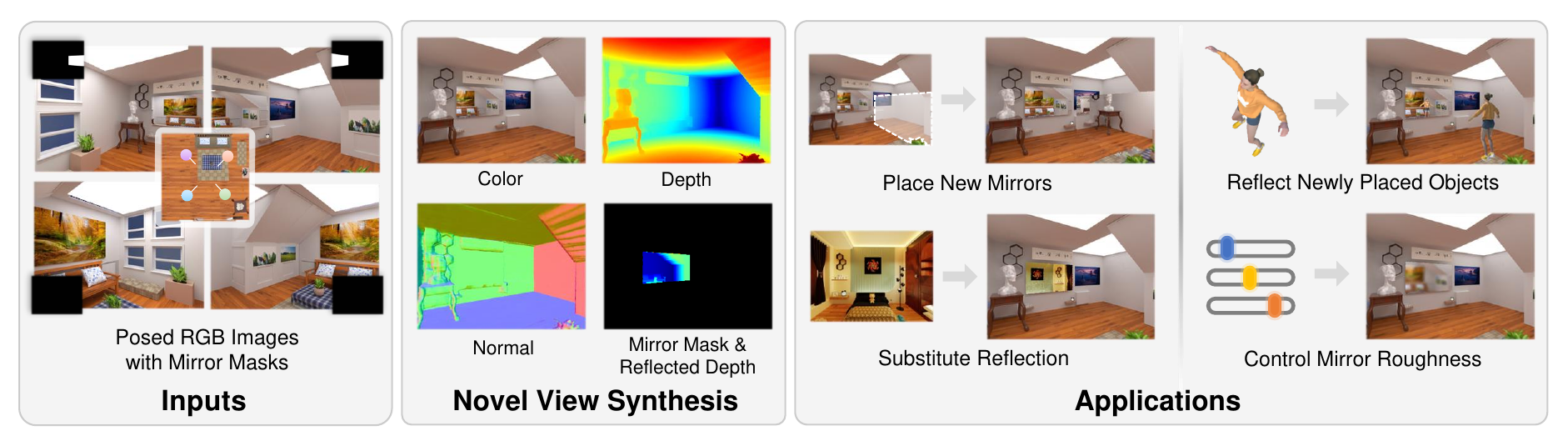}
  \vspace{-3.0em}
  \caption{
  We present Mirror-NeRF, a novel neural rendering framework that incorporates Whitted Ray Tracing to achieve photo-realistic novel view synthesis in the scene with the mirror and supports various scene manipulation applications. 
  Given the posed images with mirror reflection masks, we can learn the correct geometry and reflection of the mirror.
  }
  \Description{
  We present Mirror-NeRF, a novel neural rendering framework that incorporates Whitted Ray Tracing to achieve photo-realistic novel view synthesis in the scene with the mirror and supports various scene manipulation applications including placing new mirrors into the scene, reflecting newly placed objects, controlling mirror roughness and reflection substitution. 
  Given the posed images with mirror reflection masks, we can learn the correct geometry and reflection of the mirror.
  }
  \label{fig:teaser}
\end{teaserfigure}

\maketitle

%%%%%%%%% BODY TEXT

\section{Introduction}
\label{sec:intro}

3D scene reconstruction and rendering is a long-standing problem in the fields of computer vision and graphics with broad applications in VR and AR. Although significant progress has been made over decades, it is still very challenging to reconstruct and re-render the scenes with mirrors, which exist ubiquitously in the real world.
The "appearance" of the mirror is not multi-view consistent and changes considerably with the observer's perspective due to the physical reflection phenomenon where the light will be entirely reflected along the symmetric direction at the mirror.

\begin{figure}[t]
    \vspace{-0.5em}
    \centering
    \includegraphics[width=0.9\linewidth, trim={0 0 0 0}, clip]{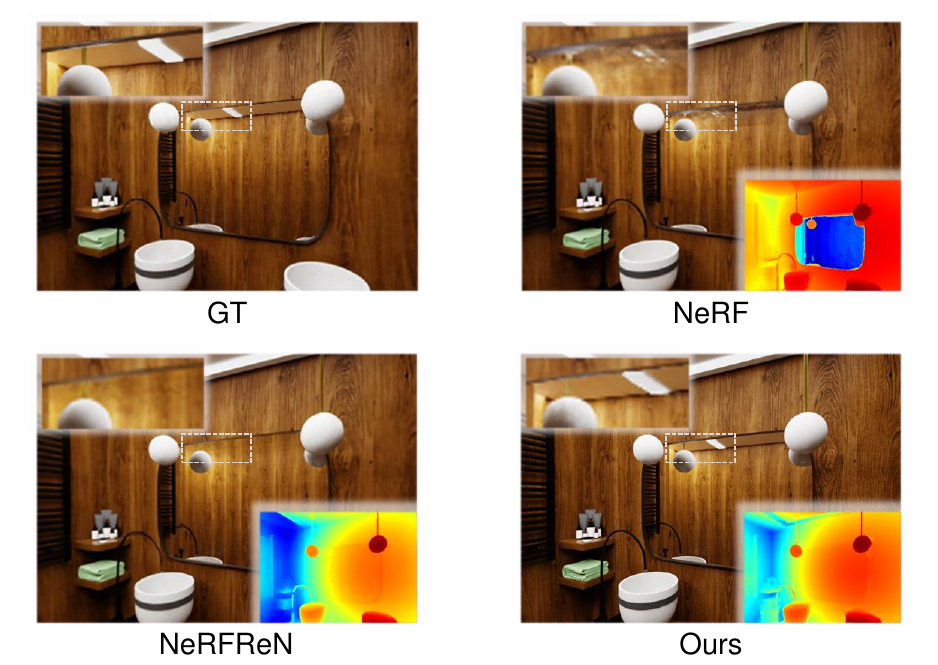}
    \vspace{-1.0em}
    \caption{
Comparison of the novel views synthesized by different methods.
NeRF~\cite{nerf} mistakes the reflection in the mirror as a separate virtual scene, leading to inaccurate depth of the mirror. NeRFReN~\cite{guo2022nerfren} uses two radiance fields to learn the color inside and outside the mirror separately. 
They synthesize the reflection in the mirror by interpolating the memorized reflection and cannot infer the reflection unobserved in the training views, \eg, the missing ceiling. 
Instead, we successfully synthesize new reflections in the mirror with the accurate depth of the mirror due to our ray tracing pipeline.
    }
    \Description{
    This figure shows the synthesized novel views on the scene with the mirror by NeRF, NerRFReN and our method. 
    NeRF and NerRFReN cannot infer the reflection unobserved in the training views, \eg, the missing ceiling, while our method successfully synthesizes new reflections in the mirror with the accurate depth of the mirror due to our ray tracing pipeline.
    }
    \label{fig:motivation}
    \vspace{-3em}
\end{figure}

Recently, Neural Radiance Fields (NeRF)~\cite{nerf} has exhibited significant success in novel view synthesis and surface reconstruction due to its capability of modeling view-dependent appearance changes. 
However, since the physical reflection is not considered in its rendering pipeline, NeRF mistakes the reflection in the mirror as a separate virtual scene, leading to the inaccurate reconstruction of the geometry of the mirror, as illustrated in Fig.~\ref{fig:motivation}.
The rendered "appearance" of the mirror also suffers from multi-view inconsistency. 
Several techniques~\cite{verbin2021ref, zhang2022modeling, srinivasan2021nerv}
decompose the object material and illuminations to model the reflection effect at the surface, but they all assume the surfaces with certain diffuse reflection to recover object surface first and then model the specular component. Thus they struggle to handle the mirrors with pure specular reflection due to the incorrect surface estimation of mirrors. 
NeRFReN~\cite{guo2022nerfren} models reflection by separating the reflected and transmitted parts of a scene as two radiance fields and improves the rendering quality for the scenes with mirrors, while it still fails to model the physical specular reflection process. Thus, it cannot render the reflection that is not observed in the training views as shown in Fig.\ref{fig:motivation}, and cannot synthesize new reflections of the objects or mirrors that are newly placed in the scene.

In this paper, we propose a novel neural rendering framework, named Mirror-NeRF, to accomplish high-fidelity novel view synthesis in the scene with mirrors and support multiple scene manipulation applications.
For clarity, we term the ray as the inverse of light.
The rays emitted from the camera are termed as camera rays and rays reflected at the surface are termed as reflected rays.
Exhaustively conducting ray tracing in a room-scale environment is prohibitively expensive. 
With the goal of achieving physically-accurate rendering of reflections in the mirror, we draw inspiration from Whitted Ray Tracing~\cite{whitted2005improved} where the ray is reflected at the mirror-like surface and terminates at a diffuse surface.
Specifically speaking, 
we first define the probability that the ray is reflected when hitting a spatial point as the reflection probability.
The reflection probability is parameterized as a continuous function in the spatial space by a Multi-Layer Perceptron (MLP).
Then we trace the ray emitted from the camera.
The physical reflection will take place when the ray hits the surface with a high reflection probability.
We accumulate the density and radiance of the ray by the volume rendering technique and synthesize the image by blending the color of camera rays and reflected rays based on the reflection probability. 
Instead of taking the specular reflection as separate neural fields, 
our neural fields are unified, which is more reasonable to synthesize new physically sound reflection from novel viewpoints.
As shown in Fig.~\ref{fig:teaser}, our representation further supports various types of scene manipulations, \eg, adding new objects or mirrors into the scene and synthesizing the reflections of these new objects in mirrors, controlling the roughness of mirrors and reflection substitution.

However, learning both geometry- and reflection-accurate mirror with the proposed new representation is not trivial. 
First, the reflection at a surface point is related to the surface normal.
The analytical surface normal from the gradient of volume density has significant noise since the density cannot concentrate precisely on the surface.
Thus, we exploit an MLP to parameterize a smooth distribution of surface normal.
Second, the reconstruction of mirror surface is ambiguous and challenging, since the "appearance" of mirror is from other objects and  not consistent from different viewpoints. Based on the fact that mirrors in real world usually have planar surfaces, we leverage both plane consistency and forward-facing normal constraints in a joint optimization manner to guarantee the smoothness of the mirror geometry and reduce the ambiguity of the reflection. Moreover, a progressive training strategy is proposed to stabilize the geometry optimization of the mirror.

\begin{figure*}
    \centering
    \vspace{-1.6em}
    \includegraphics[width=1.0\linewidth, trim={0 0 0 0.2cm}, clip]{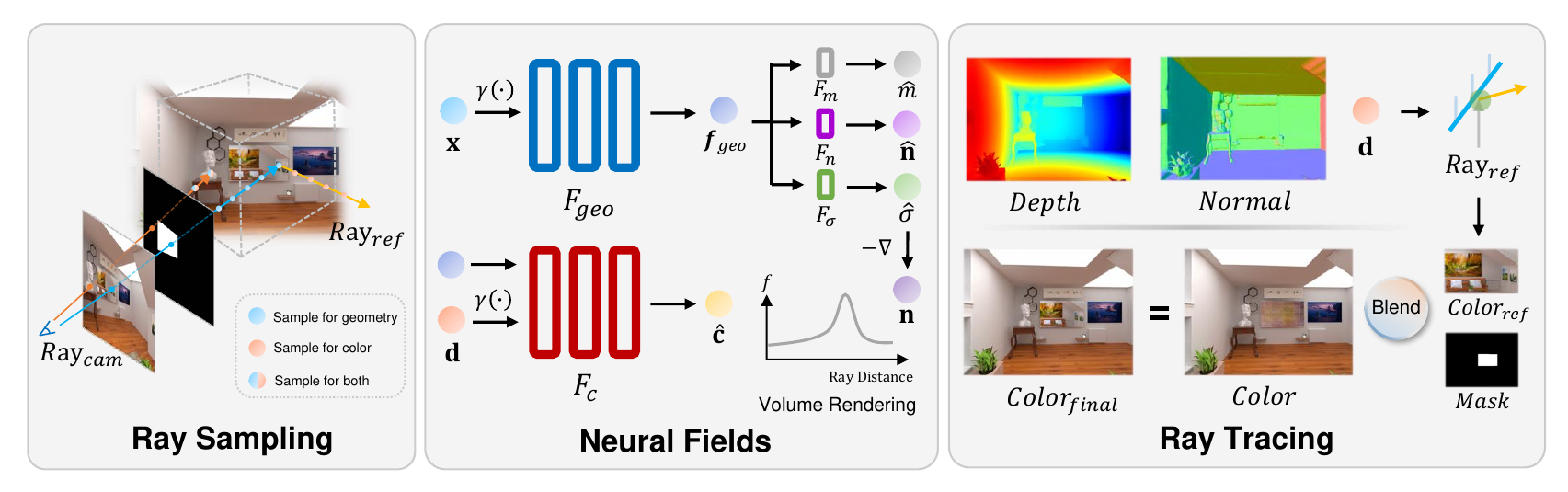}
    \vspace{-3.0em}
    \caption{
Framework.
We trace the rays physically in the scene and learn a unified radiance field of the scene with the mirror. 
The neural field takes as input spatial location $\mathbf{x}$, view direction $\mathbf{d}$, and outputs the volume density $\hat{\sigma}$, radiance $\hat{\mathbf{c}}$, surface normal $\hat{\mathbf{n}}$ and reflection probability $\hat{m}$. 
The final color is blended by the color of the camera ray and the reflected ray based on the reflection probability.
    }
    \Description{
    Unified neural fields are designed to learn the volume density, normal, reflection probability and radiance inside and outside the mirror.
    For camera rays hitting the mirror, the corresponding reflected rays are traced according to the depth and normal of the surface.
    Final pixel color is obtained by blending the volume-rendered color of camera ray and reflected ray with volume-rendered reflection probability.
    }
    \label{fig:pipeline}
    \vspace{-1.0em}
\end{figure*}

Our contributions can be summarized as follows.
\textbf{1)} We propose a novel neural rendering framework, named Mirror-NeRF, that resolves the challenge of novel view synthesis in the scene with mirrors.
Different from NeRF~\cite{nerf} and NeRFReN~\cite{guo2022nerfren} that tend to learn a separate virtual world in the mirror, Mirror-NeRF can correctly render the reflection in the mirror in a unified radiance field by introducing the reflection probability and
tracing the rays following the light transport model of Whitted Ray Tracing~\cite{whitted2005improved}. 
The physically-inspired rendering pipeline facilitates high-fidelity novel view synthesis with accurate geometry and reflection of the mirror.
\textbf{2)} To learn both accurate geometry and reflection of the mirror, we leverage several techniques, including a surface normal parametrization to acquire smooth distribution of surface normal, the plane consistency and forward-facing normal constraints with joint optimization to ensure the accurate geometry of the mirror, and a progressive training strategy to maintain the stability of training.
\textbf{3)} The proposed Mirror-NeRF enables a series of new scene manipulation applications with mirrors as shown in Fig.~\ref{fig:teaser}, such as object placement, mirror roughness control, reflection substitution, \etc.
Extensive experiments on real and synthetic datasets demonstrate that Mirror-NeRF can achieve photo-realistic novel view synthesis. A large number of scene manipulation cases show the physical correctness and flexibility of the proposed method.

\section{Related Work}
\label{sec:related_work}

\subsection{Neural Rendering}
\label{subsec:neural_radiance_fields}
The goal of neural rendering is to synthesize photorealistic images or videos by computing the light transport in a 3D scene.
Lots of works~\cite{mildenhall2019local,sitzmann2020implicit,yariv2020multiview} have been proposed to push the envelope of rendering quality in this field.
One of the most notable approaches is NeRF~\cite{nerf}, which models the radiance field of a scene using the MLP.
By training on a set of posed images, NeRF learns to infer the radiance and density of each sampled point and accumulates them along the ray with volume rendering techniques to render the color.
This enables NeRF to generate photorealistic images of the scene from a novel viewpoint.
Several extensions and improvements have been proposed to apply NeRF to more challenging problems, such as scene reconstruction~\cite{wang2021neus, yu2022monosdf, li2022vox, wu2022voxurf, wang2022go, wei2021nerfingmvs, zhang2022nerfusion, guo2022neural, azinovic2022neural, wang2022nerf, wu2022dof}, generalization~\cite{wang2021ibrnet,suhail2022generalizable}, novel view extrapolation~\cite{zhang2022ray,warburg2023nerfbusters}, scene manipulation~\cite{yang2021learning, yang2022neural, wang2022dm, yang2022neumesh, bao2023sine}, SLAM~\cite{sucar2021imap, zhu2022nice}, segmentation~\cite{zhi2021place, siddiqui2022panoptic}, human body~\cite{wang2021ibutter, peng2021neural} and so on.
Furthermore, some NeRF-variants provide various applications, such as supersampling~\cite{wang2022nerf} and controllable depth-of-field rendering~\cite{wu2022dof}.
However, these NeRF-variants struggle to model mirror reflection since they assume that all lights in the scene are reflected at Lambertain surfaces.

\subsection{Neural Rendering With Reflection}
\label{subsec:neural_radiance_fields_with_reflection}

Plenty of works ~\cite{zhang2021nerfactor,bi2020neural,jin2023tensoir,munkberg2022extracting,boss2022samurai,boss2021neural,kuang2022neroic,zhao2022factorized,zhang2023nemf} have been working on making NeRF understand physical reflection.
PhySG~\cite{zhang2021physg} simplifies light transport by modeling the environment illumination and material properties as mixtures of spherical Gaussians and integrating the incoming light over the hemisphere of the surface. 
InvRender~\cite{zhang2022modeling} extends PhySG to model the indirect light by using another mixture of spherical Gaussians to cache the light that bounces off from other surfaces.
These approaches assume that surfaces are diffuse with a simple BRDF and environment lighting is far away from the scene.
For a room with the mirror, they cannot handle the complex reflection and material diversity in the scene.
As for NeRF, it will treat the reflection in mirrors as real geometry, which reconstructs the inaccurate depth of the mirror.
RefNeRF~\cite{verbin2021ref} decomposes the light as diffuse and specular components and learns the reflection using a radiance field conditioned by the reflected view direction.
NeRFReN~\cite{guo2022nerfren} employs two radiance fields to learn the color inside and outside the mirror and depth constraints to recover the depth of the mirror.
However, these methods generate mirror reflection from new viewpoints by interpolating the previously learned reflections, and are limited in accurately inferring reflections that were not observed during training and synthesizing reflections for newly added objects or mirrors in the scene.
By introducing the physical ray tracing into the neural rendering pipeline, our method can correctly render the reflection in the mirror and support multiple scene manipulation applications.

\section{Mirror-NeRF}
\label{sec:method}

We introduce Mirror-NeRF, a physically inspired neural rendering framework that supports photo-realistic novel view synthesis of scenes with mirrors and reconstructs the accurate geometry and reflection of mirrors.
As illustrated in Fig.~\ref{fig:pipeline}, we leverage unified neural fields to learn the volume density, normal, reflection probability and radiance inside and outside the mirror (Sec.~\ref{subsec:unified_nerf}).
With the intention of generating physically-accurate reflections in the mirror, we employ the light transport model in Whitted Ray Tracing~\cite{whitted2005improved} and trace the volume rendered ray in the scene (Sec.~\ref{subsec:ray_tracing}).
Besides, some regularization constraints for the mirror surface (Sec.~\ref{subsec:regularization}) and a progressive training strategy (Sec.~\ref{subsec:training_strategy}) are proposed to improve the reconstruction quality of the mirror and stabilize the training.

\subsection{Unified Neural Fields}
\label{subsec:unified_nerf}
We design several neural fields to learn the properties of the scene, which are unified for parts inside and outside the mirror (Fig.~\ref{fig:pipeline}).

\subsubsection{Geometry and Color.}
Following the implicit representation in NeRF~\cite{nerf}, we use a geometry MLP $\mathcal{F}_{geo}$ to encode the geometry feature $f_{geo}$ at an arbitrary spatial location $\mathbf{x}$.
The volume density field is presented by a volume density MLP $\mathcal{F}_{\sigma}$ which takes $f_{geo}$ as input, and the radiance field is presented by a radiance MLP $\mathcal{F}_c$ which takes $f_{geo}$ and view direction $\mathbf{d}$ as input:
\begin{equation}
\label{eq:density_network}
\begin{split}
    & f_{geo} = \mathcal{F}_{geo} ( \gamma_x(\mathbf{x}) ), \\
    & \sigma = \mathcal{F}_{\sigma} ( f_{geo} ), \\
    & \mathbf{c} = \mathcal{F}_{c} ( f_{geo}, \gamma_d(\mathbf{d}) ),
\end{split}
\end{equation}
where $\gamma_x(\cdot)$ and $\gamma_d(\cdot)$ are respectively the positional encoding function of spatial position and view direction. $\sigma$ and $\mathbf{c}$ are volume density and radiance respectively.
To render an image from a specific viewpoint, we follow the volume rendering techniques in NeRF. 
The volume-rendered color $\hat{C}$ of a ray $\bm{r}$ is calculated by accumulating the volume densities $\sigma_i$ and radiances ${\mathbf{c}}_i$ of sampled points $x_i$ along the ray:
\begin{equation}
\label{eq:T_i_and_alpha_i}
\begin{split}
    &\hat{C}(\bm{r}) = \sum_{i=1}^{N} T_i \alpha_i {\mathbf{c}}_i, \\
    & T_i = \exp{\left(-\sum_{j=1}^{i-1}{\sigma}_j \delta_j\right)}, \\
    & \alpha_i = 1-\exp{\left(-{{\sigma}}_i \delta_i\right)},
\end{split}
\end{equation}
where $N$ is the number of sampled points on the ray $\bm{r}$, and $\delta_i$ is the sampling distance between adjacent points along the ray.

\subsubsection{Smooth Surface Normal.}
Prior works~\cite{srinivasan2021nerv,boss2021nerd} have analyzed the acquisition of surface normal in NeRF that the negative gradient of volume density \wrt~$\mathbf{x}$ can give a differentiable approximation of the true normal:
\begin{equation}
\label{eq:normal_calculation}
    \mathbf{n} = - \frac{\nabla \sigma(\mathbf{x})}{|| \nabla \sigma(\mathbf{x}) ||}.
\end{equation}
However, such parametrization tends to produce an unsmooth surface normal distribution since the volume density cannot concentrate precisely on the surface.
The noise in the surface normal will severely hamper tracing the correct direction of the reflected rays at the mirror.
To obtain a smooth distribution of surface normal, we utilize an MLP $\mathcal{F}_{n}$ that takes $f_{geo}$ as input and predicts the smoothed surface normal $\hat{\mathbf{n}}$:
\begin{equation}
\label{eq:normal_network}
    \hat{\mathbf{n}} = \mathcal{F}_{n} ( f_{geo} ).
\end{equation}
We supervise the optimization of $\mathcal{F}_{n}$ by the analytical surface normal $\mathbf{n}$:
\begin{equation}
\label{eq:normal_supervision}
    \mathcal{L}_{n} = || \hat{\mathbf{n}} - \mathbf{n} ||^2_2.
\end{equation}
To compute the surface normal at the intersection point of a ray $\bm{r}$ and the surface, we follow the Eq.~(\ref{eq:T_i_and_alpha_i}) by:
\begin{equation}
\label{eq:normal_surface}
    \hat{\mathbf{N}}(\bm{r}) = \sum_{i=1}^{N} T_i \alpha_i \hat{\mathbf{n}}_i .
\end{equation}

\subsubsection{Reflection Probability.}
To model the reflection and perform the Whitted-style ray tracing described in Sec.~\ref{subsec:ray_tracing}, we also utilize an MLP $\mathcal{F}_{m}$ to predict the probability $m$ that rays will be reflected at a spatial point:
\begin{equation}
\label{eq:mirror_mask_network}
    m = \mathcal{F}_{m} ( f_{geo} ),
\end{equation}
where $m$ ranges in $[0,1]$.
To determine the reflection probability $\hat{M}$ of a ray $\bm{r}$ hitting the solid surface, we perform the volume rendering like Eq.~(\ref{eq:T_i_and_alpha_i}):
\begin{equation}
\label{eq:mirror_mask}
    \hat{M}(\bm{r}) = \sum_{i=1}^{N} T_i \alpha_i m_i.
\end{equation}

\begin{figure}[t]
    \centering
    \includegraphics[width=0.9\linewidth, trim={0 0 0 0}, clip]{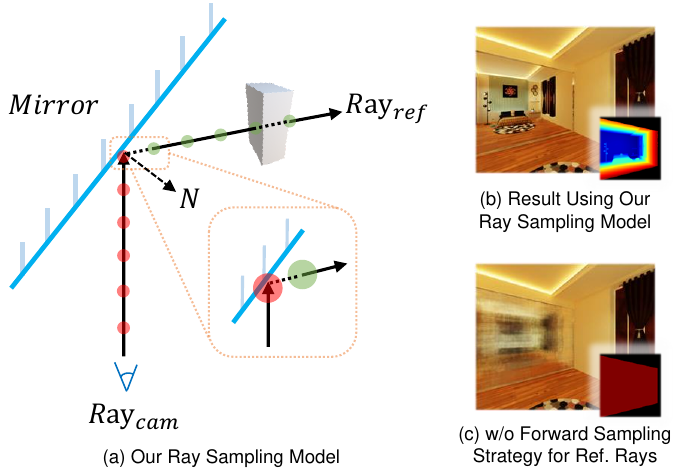}
    \vspace{-1.0em}
    \caption{
Our strategy for sampling points on rays is shown in (a).
We sample points on both the camera ray and the reflected ray.
For the reflected ray, we forward a distance from the origin to start sampling points to avoid the reflected ray terminating unexpectedly near the origin due to the "foggy" geometry. 
The effectiveness of this design is demonstrated by the comparison of (b) and (c) where mirror reflection is corrupted without the forward sampling strategy. 
The bottom right images in (b) and (c) show the reflected depth of the mirror.
    }
    \Description{
    For camera rays hitting the mirror, we sample points on both the camera ray and the reflected ray.
    The geometry seen from camera is volume rendered by the sampled points on the camera ray, while the color seen from camera is combined by the volume-rendered color of both the camera ray and the reflected ray.
    For the reflected ray, we start sampling after skipping a distance from the origin to avoid the reflected ray terminating unexpectedly near the origin due to the "foggy" geometry near the mirror induced by the density-based representation.
    }
    \label{fig:sampling}
    \vspace{-1.7em}
\end{figure}

\subsection{Whitted-Style Ray Tracing}
\label{subsec:ray_tracing}

NeRF~\cite{nerf} does not take into account the physical reflection in the rendering pipeline.
When applied to the scene with the mirror, NeRF cannot reconstruct the geometry of the mirror and treats the reflection in the mirror as a separate virtual scene.
To handle the reflection at the mirror, we draw inspiration from Whitted Ray Tracing~\cite{whitted2005improved} where the ray is reflected at the mirror-like surface and terminates at the diffuse surface.
As shown in Fig.~\ref{fig:sampling}, when a ray is reflected, we first compute the location $\hat{\mathbf{X}}$ of the intersection point of the ray $\bm{r}$ and the surface by:
\begin{equation}
\label{eq:x_surf}
\begin{split}
    & \hat{\mathbf{X}}(\bm{r}) = \mathbf{o}(\bm{r}) + \hat{D}(\bm{r}) \mathbf{d}(\bm{r}), \\
    & \hat{D}(\bm{r}) = \sum_{i=1}^{N} T_i \alpha_i t_i,
\end{split}
\end{equation}
where $\hat{D}$, $\mathbf{o}$ and $\mathbf{d}$ are the expected termination depth, origin and direction of the ray $\bm{r}$ respectively. $T_i$ and $\alpha_i$ are the same as Eq.~(\ref{eq:T_i_and_alpha_i}).

To trace the reflected ray $\bm{r}_{ref}$ of a ray $\bm{r}$, we set $\hat{\mathbf{X}}(\bm{r})$ as its origin, and compute its direction by:
\begin{equation}
\label{eq:dir_ref}
\begin{split}
    & \mathbf{d}(\bm{r}_{ref}) = \mathbf{d}(\bm{r}) - 2 \left( \hat{\mathbf{N}}(\bm{r}) \cdot \mathbf{d}(\bm{r}) \right) \hat{\mathbf{N}}(\bm{r}).
\end{split}
\end{equation}
Here all direction vectors are normalized.

Then, we use the volume rendering technique to compute the color of the ray $\bm{r}$ and its reflected ray $\bm{r}_{ref}$. 
The radiances of the sampled points on $\bm{r}$ and $\bm{r}_{ref}$ are attained by querying the same neural radiance field.
Since the density-based representation always induces a "foggy" geometry, the reflected ray may terminate unexpectedly near the origin as illustrated in Fig.~\ref{fig:sampling}(c).
To solve the problem, we start sampling points on the reflected ray at a distance from the origin as shown in Fig.~\ref{fig:sampling}(a).

We blend the color of the ray $\bm{r}$ and its reflected ray $\bm{r}_{ref}$ according to the volume-rendered reflection probability of the ray $\hat{M}(\bm{r})$ as:
\begin{equation}
\label{eq:color_blend}
    \hat{C^P}(\bm{r}) = \hat{C}(\bm{r}) \left( 1 - \hat{M}(\bm{r}) \right) + \hat{C^P}(\bm{r}_{ref}) \hat{M}(\bm{r}).
\end{equation}
Note that $\hat{C^P}$ is defined recursively, and the recursion terminates when $\hat{M}$ is zero or the specified maximum recursion depth is reached.

For each pixel, we generate a ray from the camera and trace it in the scene.
The set of these camera rays is denoted as $R_{cam}$.
The pixel color is rendered by Eq.(\ref{eq:color_blend}) with $\bm{r} \in R_{cam}$.
We supervise the rendered pixel color by the ground truth pixel color $C^I$ with a photometric loss: 
\begin{equation}
\label{eq:color_loss}
    \mathcal{L}_{c} =  \sum_{\bm{r} \in R_{cam}} ||\hat{C^P}(\bm{r}) - C^I(\bm{r})||^2_2.
\end{equation}

To guide the optimization of reflection probability $\hat{M}$, we calculate the binary cross entropy loss between the rendered reflection probability $\hat{M}$ and the mirror reflection mask $M$:
\begin{equation}
\label{eq:mask_loss}
\begin{split}
    &\mathcal{L}_{m} =  \sum_{r \in R_{cam}} - \left( M(\bm{r}) \log{\hat{M}(\bm{r})} + \left(1-M(\bm{r})\right) \log{\left(1 - \hat{M}(\bm{r})\right)} \right),
\end{split}
\end{equation}
where $M$ is obtained by using the off-the-shelf segmentation tools like ~\cite{kirillov2023segment} on the ground-truth images.

\subsection{Regularization}
\label{subsec:regularization}

We design a novel rendering pipeline based on Whitted Ray Tracing for the mirror, while a na\"ive training without regularization always leads to unstable convergence at the mirror where the "appearance" of the mirror is blurred.
We find that the bumpy surface of the mirror will greatly affect the quality of reflection due to underconstrained density at the mirror.
Thus, we introduce several regularization terms into our optimization process.

\subsubsection{Plane Consistency Constraint.} 
As far as we observe, mirrors typically have planar surfaces in the real world.
To make full use of this property, we apply the plane consistency constraint proposed by ~\cite{chen2022structnerf} to the surface of the mirror.
Specifically, we randomly sample four points $A$, $B$, $C$, $D$ on the surface of the mirror and enforce the normal vector of the plane $ABC$ to be perpendicular to the vector $\vec{AD}$:
\begin{equation}
\label{eq:plane_consistency}
    \mathcal{L}_{pc} = \frac{1}{N_p} \sum_{i=1}^{N_p} | \vec{A_iB_i} \times \vec{A_iC_i} \cdot \vec{A_iD_i} | ,
\end{equation}
where $N_p$ denotes the number of the 4-point sets randomly selected from the planes.

\subsubsection{Forward-facing Normal Constraint.}
With regard to the reflection equation Eq.~(\ref{eq:dir_ref}), we find that it still holds when the surface normal rotates 180 degrees and points to the inside of the surface.
This ambiguity will incur the incorrect depth of the mirror.
To tackle this issue, we follow ~\cite{verbin2021ref} to enforce that the analytical surface normal $\hat{\mathbf{n}}$ of sampled points makes an obtuse angle with the direction $\mathbf{d}$ of the camera ray $\bm{r}$, \ie, the surface normal should be forward-facing to the camera. 
\begin{equation}
\label{eq:normal_reg}
    \mathcal{L}_{n_{reg}} = \max(0, \hat{\mathbf{n}} \cdot \mathbf{d}(\bm{r}))^2.
\end{equation}

\subsubsection{Joint Optimization.}
In practice, we jointly optimize all networks with the aforementioned losses.
In other words, each loss will eventually have an impact on the volume density field and radiance field:
\begin{equation}
\label{eq:total_loss}
\begin{split}
    & \mathcal{L} = \lambda_{c} \mathcal{L}_{c} + \lambda_{m} \mathcal{L}_{m} + \lambda_{pc} \mathcal{L}_{pc} \\
    & \qquad + \lambda_{n} \mathcal{L}_{n} + \lambda_{n_{reg}} \mathcal{L}_{n_{reg}},
\end{split}
\end{equation}
where $\lambda$ is the coefficient of each loss term.
Joint optimization will bring three main advantages.
First, the surface normal loss $\mathcal{L}_{n}$ not only influences the $\mathcal{F}_{n}$ but also encourages $\mathcal{F}_{geo}$ to produce a smooth feature distribution, which makes the volume density uniformly concentrate on the surface to strengthen the flatness of the surface. 
Second, the reflection probability loss $\mathcal{L}_{m}$ will promote the volume density field to reach a peak at the mirror, thereby producing an unbiased depth for the mirror. 
Both of the losses regulate the $\mathcal{F}_{geo}$ through $f_{geo}$.
Third, in spite of the employment of plane and normal constraints, any tiny error of the surface normal will be amplified during the reflection.
Through joint optimization, these errors will be iteratively refined since the photometric loss $\mathcal{L}_{c}$ will implicitly adjust the surface normal $\hat{\mathbf{N}}$ to the desired direction through the differentiable reflection equation.

\subsection{Progressive Training Strategy}
\label{subsec:training_strategy}
In the early stage of training, the neural field is unstable and easily falls into the local optimum.
We conclude the degeneration situations as two cases:
1) The reflection in the mirror might be learned as a separate scene with inaccurate depth just like NeRF in the case the color converges faster than the geometry.
2) The color may be stuck in a local optimum and blurry if strong geometric regularization is enabled at the beginning.
To make training stable, we progressively train the image area inside and outside the mirror and schedule the coefficients of losses at different stages of training.
In the initial stage, we enable $\lambda_{c}$ and disable the remaining coefficients to maintain the stability of the neural field and avoid the geometry of the mirror being ruined.
Furthermore, we replace the $\mathcal{L}_{c}$ with masked photometric loss $\mathcal{L}_{cm}$:
\begin{equation}
\label{eq:color_loss_masked}
    \mathcal{L}_{cm} =  \sum_{\bm{r} \in R_{cam} \bigcap \overline{R_M}}  ||\hat{C^P}(\bm{r}) - C^I(\bm{r})||^2_2 + \sum_{\bm{r}\in R_{cam} \bigcap R_M} ||\hat{C^P}(\bm{r}) - K||^2_2, \\
\end{equation}
where $R_M$ is the set of rays hitting the mirror-like surface
and $\overline{R_M}$ is the complementary set of $R_M$. $K$ is a constant vector, which we use $(0,0,0)$ in our experiments.
The use of $K$ for the image region inside the mirror is to learn an initial rough shape of the mirror without learning its reflection, which will be discussed in Sec.~\ref{subsubsec:masked_rgb_loss}.
$\mathcal{L}_{cm}$ is used until the last stage.
After a few epochs, we activate the $\lambda_{m}$, $\lambda_{pc}$, $\lambda_{n}$, $\lambda_{n_{reg}}$ to regularize the location and geometry of the mirror.
After this stage, the accurate depth of the mirror is expected to have been learned by the neural fields.
At last, we use $\mathcal{L}_{c}$ instead of $\mathcal{L}_{cm}$ to jointly optimize the reflection part and refine the geometry of the mirror.

\begin{figure*}%[!htbp]
    \vspace{-1em}
    \centering
    \includegraphics[width=1.0\linewidth, trim={0 0 0 0}, clip]{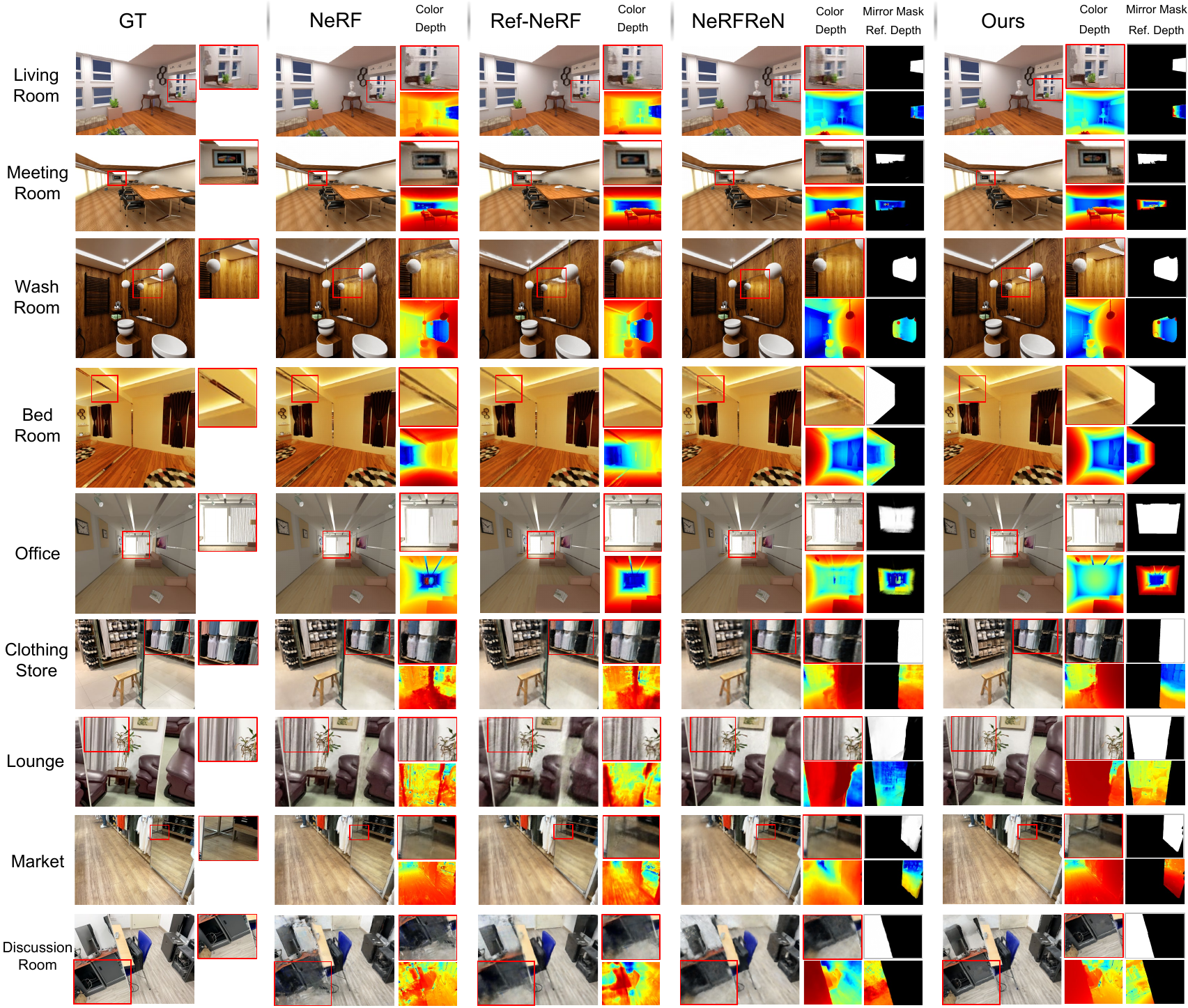}
    \vspace{-2.0em}
    \caption{
Qualitative comparison of novel view synthesis on synthetic and real scenes with mirrors.
    }
    \Description{
    Qualitative comparison of NeRF~\cite{nerf}, Ref-NeRF~\cite{verbin2021ref}, NeRFReN~\cite{guo2022nerfren} and our method on novel view synthesis on synthetic and real scenes with mirrors.
    }
    \vspace{-1.5em}
    \label{fig:comparisons}
\end{figure*}

\section{Experiments}
\label{sec:experiments}

\subsection{Datasets}
\label{subsec:data}
To the best of our knowledge, there is no room-level dataset containing mirrors publicly available for the task of novel view synthesis.
Therefore, we create 5 synthetic datasets and capture 4 real datasets with mirrors.
Each synthetic dataset is an indoor room downloaded from the BlendSwap~\cite{blenderswap}, including living room, meeting room, washroom, bedroom, and office.
Real datasets are captured in real indoor scenes using IPad Pro, including clothing store, lounge, market and discussion room.
In each dataset, images are captured 360 degrees around the scene.
We split the images as training and test sets to perform the quantitative and qualitative comparison.
We use the off-the-shelf segmentation tool ~\cite{kirillov2023segment} to segment the mirror reflection mask in the image.

\subsection{Comparisons}
\label{subsec:comparisons}

We compare our method with NeRF~\cite{nerf} and the state-of-the-art neural rendering methods dealing with the reflection, \ie, Ref-NeRF~\cite{verbin2021ref} and NeRFReN~\cite{guo2022nerfren}. 
The same mirror masks are provided for our method and NeRFReN.

We perform the quantitative comparisons of novel view synthesis on the metrics PSNR, SSIM~\cite{wang2004image}, and LPIPS~\cite{zhang2018unreasonable}.
As demonstrated in Tab.~\ref{tab:avg_compare}, on the regular test viewpoints, our method outperforms the SOTA methods handling the reflection (\ie, Ref-NeRF and NeRFReN) on both synthetic and real datasets, and is comparable with NeRF.
Note that NeRF does not reconstruct the physically sound geometry of the mirror and just interpolates the memorized reflection when performing novel view synthesis,
while our method recovers the correct depth of the mirror and enables synthesizing reflections unobserved in training views and multiple applications due to the physical ray-tracing pipeline.
Since the above test viewpoints are close to the distribution of training viewpoints, NeRF can generate visually reasonable reflection by interpolating the reflection of nearby views.
To compare the correctness of modeling reflection, we capture a set of more challenging test images with more reflections unobserved in the training views.
We quantitatively compare the reflection in the mirror, as shown in Tab.~\ref{tab:avg_mirror_compare}.
Our method surpasses all the compared methods since we can faithfully synthesize the reflection by tracing the reflected ray in the scene.
Please refer to the supplementary material for more details.

Qualitative comparisons on the synthetic and real datasets are shown in Fig.~\ref{fig:comparisons}.
NeRF models the scene as a volume of particles that block and emit light~\cite{tewari2022advances}, and conditions the view-dependent reflection by view direction input.
The assumption is suitable well for the Lambertian surface but fails in resolving the reflection in the mirror. 
The multi-view inconsistent reflection in the mirror will mislead NeRF to learn a separate virtual scene in the mirror, \eg, the inaccurate depth results shown in Fig.~\ref{fig:comparisons}, since NeRF does not consider the physical reflection in the rendering pipeline.
Despite Ref-NeRF's attempt to reproduce reflections by reparameterizing the radiance field using the reflected ray direction and surface materials, it encounters the same limitation as NeRF in reconstructing the mirror's geometry.
NeRFReN takes two neural radiance fields to model the scene inside and outside the mirror respectively and can produce the smooth depth of the mirror.
However, the above methods synthesize the reflection by interpolating the memorized reflection.
The common drawback of these methods is that they cannot synthesize the reflections unobserved in the training set from new viewpoints, \eg, the missing statue in the mirror of the living room, the vanishing ceiling in the mirror of the washroom, and broken cabinet in the mirror of the discussion room in Fig.\ref{fig:comparisons}.
With our neural rendering framework based on physical ray tracing, we can synthesize the reflection of any objects in the scene from arbitrary viewpoints.
Moreover, NeRF, Ref-NeRF, and NeRFReN struggle to produce the reflection of the objects whose reflection has high-frequency variations in color, \eg, the distorted hanging picture in the mirror of the meeting room, the blurry curtain in the mirror of the office and the lounge, and the "fogged" clothes in the mirror of the clothing store in Fig.\ref{fig:comparisons}.
By contrast, our method renders detailed reflections of objects by tracing the reflected rays.
Compared to NeRFReN, our method can also recover smoother depth of the mirror, \eg, the depth of the mirror from NeRFReN is damaged by the reflection of distant light on the office while our method recovers the mirror depth accurately.

\begin{table}
  \centering
  \resizebox{1.0\linewidth}{!}{

  \renewcommand\tabcolsep{4pt}
  \begin{tabular}{l|cccccc}
    \toprule
        
    \multicolumn{1}{c}{\multirow{3}{*}{Methods}} & \multicolumn{3}{c}{Synthetic Datasets} & \multicolumn{3}{c}{Real Datasets}\\
    
    \cmidrule(lr){2-4} \cmidrule(lr){5-7} 

     & PSNR $\uparrow$ & SSIM $\uparrow$ & LPIPS $\downarrow$ & PSNR $\uparrow$ & SSIM $\uparrow$ & LPIPS $\downarrow$\\

    \midrule
    
    NeRF & 28.501 & 0.903 & \cellcolor{tablered}0.066 & \cellcolor{tablered}25.399 & \cellcolor{tablered}0.788 & \cellcolor{orange}0.209\\
    Ref-NeRF & \cellcolor{orange}28.703 & \cellcolor{orange}0.905 & 0.079 & 24.544 & 0.730 & 0.294\\
    NeRFReN & 28.483 & 0.902 & 0.080 & 23.191 & 0.686 & 0.367\\
    Ours & \cellcolor{tablered}29.243 & \cellcolor{tablered}0.907 & \cellcolor{orange}0.077 & \cellcolor{orange}25.173 & \cellcolor{orange}0.785 & \cellcolor{tablered}0.205\\

    \bottomrule
  \end{tabular}
  }
  \caption{
  Quantitative comparison of novel views at regular test viewpoints on synthetic and real scenes with mirrors.
  The best is marked in red and the second is marked in orange.
  }
  \label{tab:avg_compare}
  \vspace{-2em}
\end{table}

\begin{table}
  \centering
  \resizebox{1.0\linewidth}{!}{

  \renewcommand\tabcolsep{4pt}

  \begin{tabular}{l|cccccc}
    \toprule
        
    \multicolumn{1}{c}{\multirow{3}{*}{Methods}} & \multicolumn{3}{c}{Synthetic Datasets} & \multicolumn{3}{c}{Real Datasets}\\
    
    \cmidrule(lr){2-4} \cmidrule(lr){5-7} 

     & PSNR $\uparrow$ & SSIM $\uparrow$ & LPIPS $\downarrow$ & PSNR $\uparrow$ & SSIM $\uparrow$ & LPIPS $\downarrow$\\

    \midrule
    
    NeRF & 23.326 & 0.964 & \cellcolor{orange}0.027 & 19.749 & 0.886 & \cellcolor{orange}0.117 \\
    Ref-NeRF & 22.828 & 0.964 & 0.028 & \cellcolor{orange}20.188 & \cellcolor{orange}0.897 & 0.122 \\
    NeRFReN & \cellcolor{orange}23.542 & \cellcolor{orange}0.966 & 0.030 & 19.174 & 0.871 & 0.148 \\
    Ours & \cellcolor{tablered}25.677 & \cellcolor{tablered}0.975 & \cellcolor{tablered}0.021 & \cellcolor{tablered}22.705 & \cellcolor{tablered}0.912 & \cellcolor{tablered}0.085 \\

    \bottomrule
  \end{tabular}
  
  }
  \caption{
  Quantitative comparison of reflections inside the mirror from challenging novel viewpoints out of the training set distribution on synthetic and real scenes.
  }
  \label{tab:avg_mirror_compare}
  \vspace{-2em}
\end{table}

\begin{table}[tb]
\centering
\resizebox{1.0\linewidth}{!}{
\tabcolsep 10pt
\begin{tabular}{l|ccc}
\toprule
\multicolumn{1}{c|}{Settings} & PSNR $\uparrow$ & SSIM $\uparrow$ & LPIPS $\downarrow$ \\ 
\hline
w/o Surface Normal Param. & 20.464 & 0.720 & 0.349 \\
w/o $\mathcal{L}_{cm}$ & 28.331 & 0.878 & 0.103 \\
w/o Plane Consistency & 30.687 & 0.916 & 0.058 \\
w/o Forward. Normal Reg. & 31.108 & 0.923 & 0.052 \\
w/o Joint Optimization & 27.691 & 0.875 & 0.106 \\
Full Model & \cellcolor{tablered}32.422 & \cellcolor{tablered}0.933 & \cellcolor{tablered}0.047 \\
\bottomrule
\end{tabular}
}
\caption{
We quantitatively analyze our model design and training schemes on the synthetic bedroom.
}
\label{tab:ablation}
\vspace{-3.0em}
\end{table}

\begin{figure}[t]
    \centering
    \vspace{-0.5em}
    \includegraphics[width=0.95\linewidth, trim={0 0 0 0}, clip]{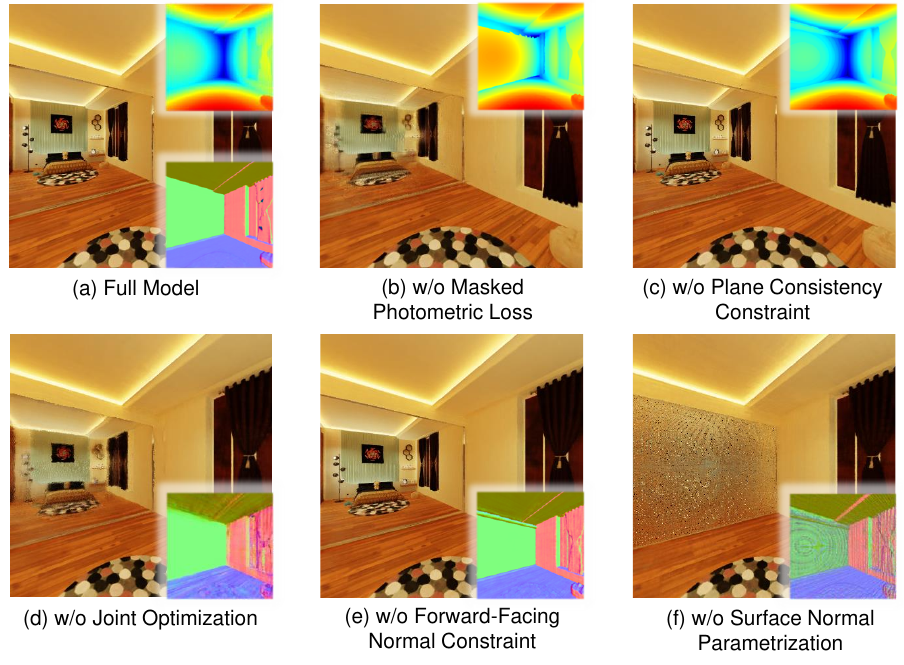}
    \vspace{-1.5em}
    \caption{
Ablation studies.
We qualitatively analyze our model design and training schemes.
The top right and bottom right images in each subfigure show the depth and normal map respectively.
    }
    \Description{
    We qualitatively analyze our model design and training schemes.
    Ablation studies includes masked photometric loss, plane consistency constraint, joint optimization, forward-facing normal constraint and surface normal parametrization.
    }
    \label{fig:ablation}
    \vspace{-1.5em}
\end{figure}

\subsection{Ablation Studies}
\label{subsec:ablation_study}

We qualitatively and quantitatively analyze our model design and training schemes on the synthetic bedroom in this section, as shown in Fig.~\ref{fig:ablation} and Tab.~\ref{tab:ablation}.
For more ablation studies, please refer to the supplementary material.

\subsubsection{Smooth Surface Normal Parametrization.}
We first inspect the effectiveness of our surface normal parametrization (Sec.\ref{subsec:unified_nerf}) by using the analytical surface normal from Eq.~(\ref{eq:normal_calculation}) to calculate the direction of the reflected ray.
As depicted in Fig.~\ref{fig:ablation}(f) and Tab.~\ref{tab:ablation}, the reflection in the mirror is collapsed due to the inevitable noise in the analytical surface normal of the mirror.
Instead, our parametrization provides a smooth surface normal with less noise to guide the optimization of the reflection in the mirror.

\subsubsection{Masked Photometric Loss $\mathcal{L}_{cm}$.}
\label{subsubsec:masked_rgb_loss}
Without the usage of $\mathcal{L}_{cm}$ in the early stage (Sec.~\ref{subsec:training_strategy}), the depth of the mirror is incorrectly recovered as depicted in Fig.~\ref{fig:ablation}(b).
The reason for this is that color supervision inside the mirror may lead to the optimization of mirror geometry getting stuck in a local optimum during the initial stages while the mirror geometry has not yet converged.

\subsubsection{Regularization.}
We then analyze the efficacy of each regularization term (Sec.\ref{subsec:regularization}) by turning it off during training.
As demonstrated in Fig.~\ref{fig:ablation}(c) and Tab.~\ref{tab:ablation}, without plane consistency constraint, the discontinuities occur in the depth of the mirror which decreases the image quality.
A similar effect happens for the forward-facing normal constraint as shown in Fig.~\ref{fig:ablation} (e).
This normal regularization can improve the image quality by correctly orienting the surface normal to the room.
Without the joint optimization strategy, the reflection in the mirror is blurred due to the imprecise geometry of the mirror as shown in Fig.~\ref{fig:ablation} (d).
When all regularization terms are enabled, we successfully learn the precise reflection in the mirror with the highest image quality.

\subsection{Applications}
\label{subsec:exp_applications}
\begin{figure*}
    \centering
    \includegraphics[width=1.0\linewidth, trim={0 0 0 0}, clip]{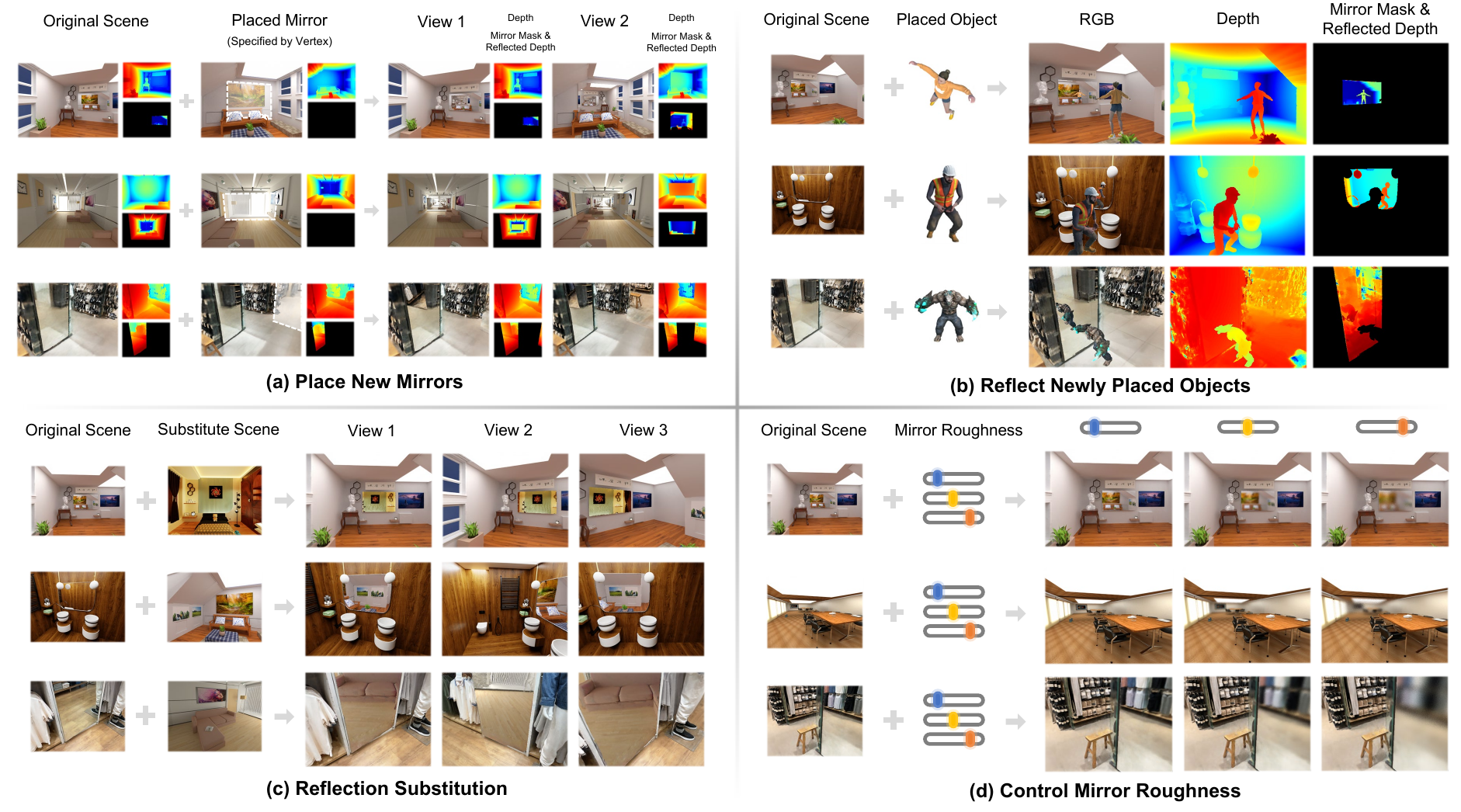}
    \vspace{-2.5em}
    \caption{
Applications on synthetic and real scenes with mirrors.
    }
    \Description{
    Our method supports various scene manipulation applications including placing new mirrors into the scene, reflecting newly placed objects, controlling mirror roughness and reflection substitution. 
    }
    \label{fig:applications}
    \vspace{-1.0em}

\end{figure*}

Due to the physical modeling of the mirror reflection, the proposed Mirror-NeRF supports various new scene manipulation applications with mirrors as shown in Fig.~\ref{fig:applications}.

\subsubsection{Placing New Mirrors.}
By tracing the reflected rays at the mirror recursively, it is feasible for our method to integrate new mirrors into the original scene. 
As shown in Fig.~\ref{fig:applications}(a), we enable the synthesis of novel views involving inter-reflection between the newly placed mirror and the original mirror, \eg, the endless reflection of the room in the new and original mirrors in the first two rows, and the new reflection of the ground in the last row.

\subsubsection{Reflecting Newly Placed Objects.}
We support the composition of multiple neural radiance fields and synthesize new reflections of the composite scenes in the mirror.
Specifically, for each traced ray, we detect occlusion by comparing the volume-rendered depth from the radiance fields that have a collision with the ray.
The ray will hit the surface with the minimum depth, and terminate or be reflected at the surface.
Here we show the composite results of dynamic radiance field D-NeRF~\cite{pumarola2021d} with the scene modeled by our method in Fig.~\ref{fig:applications}(b).
The reflection of objects from D-NeRF is precisely synthesized in the mirror.
This application might be of great use in VR and AR.
Please refer to the supplementary video for the vivid dynamic composite results.

\subsubsection{Reflection Substitution.}
In the film and gaming industries, artists may desire to create some magical visual effects, for example, substituting the reflections in the mirror with a different scene.
Since we learn the precise geometry of the mirror, it can be easily implemented by transforming the reflected rays at the mirror into another scene and rendering the results of the reflected ray.
As shown in Fig.~\ref{fig:applications}(c), we can synthesize the photo-realistic view of the new scene in the mirror with multi-view consistency.
Note that in consequence of tracing reflected rays in the new scene, the appearance in the mirror is flipped compared to the new scene.

\subsubsection{Controlling the Roughness of Mirrors.}
According to the microfacet theory \cite{walter2007microfacet}, the reason why a surface looks rough is that it consists of a multitude of microfacets facing various directions.
We support modifying the roughness of the mirror by simulating the microfacet theory.
Specifically, we trace the camera ray multiple times following Eq.\ref{eq:dir_ref} with different random noises added on the surface normal and average the volume-rendered colors to get the final color of this ray.
The roughness of the mirror is controlled by the magnitude of noise and the number of tracing times.
With this design, we can generate reasonable reflections with different roughness as shown in Fig.~\ref{fig:applications}(d).

\section{Conclusion}
\label{sec:conclusion}

We have proposed a novel neural rendering framework following Whitted Ray Tracing, which synthesizes photo-realistic novel views in the scene with the mirror and learns the accurate geometry and reflection of the mirror.
Besides, we support various scene manipulation applications with mirrors.
As a limitation, our method does not explicitly estimate the location of the light source in the room, which prevents us from relighting the room.
The refraction is also not modeled in our framework since we focus on mirrors currently, and it is naturally compatible with our ray tracing pipeline and considered as future work.

\begin{acks}
    This work was partially supported by the NSFC (No.~62102356) and Ant Group.
\end{acks}

\clearpage
\bibliographystyle{ACM-Reference-Format}
\balance
\bibliography{references}

\end{document}